\documentclass[10pt]{article} 

\usepackage[preprint]{rlj} 

\usepackage{amssymb}            
\usepackage{mathtools}          
\usepackage{mathrsfs}           
\usepackage{graphicx}           
\usepackage{subcaption}         
\usepackage[space]{grffile}     
\usepackage{url}                
\usepackage{lipsum}             
\usepackage{booktabs}
\usepackage{algorithm}
\usepackage{algorithmic}

\newcommand{\nomealgo}{ARC-RL}

\title{ARC-RL: A Reinforcement Learning Playground Inspired by ARC Raiders}

\setrunningtitle{ARC-RL: A Reinforcement Learning Playground}

\author{Carlo Romeo\textsuperscript{1}, Andrew D. Bagdanov\textsuperscript{1}}

\coverPageAuthor{Carlo Romeo, Andrew D. Bagdanov}

\emails{
\textsuperscript{1}\{carlo.romeo, andrew.bagdanov\}@unifi.it}

\affiliations{
$^{1}$\textbf{Media Integration and Communication Center -- University of Florence}
}

\contribution{
    We introduce \nomealgo, a custom suite of four continuous-control MuJoCo environments featuring robotic morphologies inspired by the bestiary of \emph{ARC Raiders}: two hexapods with three-link legs (\emph{Queen}, \emph{Tick}), one armoured hexapod with two-link legs (\emph{Bastion}), and one quadruped with three-link legs (\emph{Leaper}).
    }
    {
    Existing legged-locomotion benchmarks bundle robots derived from real commercial hardware (ANYmal, Unitree Go1, Boston Dynamics Spot, Cassie, Barkour) and therefore inherit a narrow distribution of leg counts, body plans, and proportions.
    }

\contribution{
    We provide a unified, multi-component reward structure that combines a velocity-tracking tent, a survive bonus, a phase-locked gait-compliance term, and standard regularizers and safety penalties --- the same closed-form expression across all four morphologies, with per-robot adaptation expressed only through weights and parameters.
    }
    {
    Game NPCs must avoid the erratic, ``drunken'' gaits typical of bare-bones MuJoCo-Ant-style rewards, while simultaneously satisfying stylistic constraints that have no direct counterpart in the sim-to-real robotics literature.
    }

\contribution{
    We supply hand-crafted Central Pattern Generator (CPG) demonstrators for each morphology and present a controlled empirical study of how standard online RL (SAC, SPEQ, SOPE-EO) and online-with-prior-data methods (SACfD, SPEQ-O2O, SOPE) behave on this playground.
    }
    {
    Modern reward-design recipes for legged locomotion produce competent locomotion but not necessarily \emph{stylised} locomotion; isolating how different RL paradigms cope with strict animation constraints requires a controlled, morphology-diverse benchmark.
    }

\keywords{Legged locomotion, animation-style reinforcement learning, MuJoCo benchmarks, central pattern generators, online RL with prior data}

\summary{
The history of deep reinforcement learning has been organised, to a remarkable degree, around the conquest of game benchmarks, and a parallel strand has produced increasingly sophisticated continuous-control playgrounds for legged locomotion. The transposition of these techniques to commercial games, however, introduces a distinct set of constraints: NPC motion must be visually believable at every frame, must conform to designer-specified style, and frequently inhabits morphologies with no real-robot counterpart. To investigate this regime in a reproducible setting, we introduce \textbf{ARC-RL}, a suite of four MuJoCo environments featuring robotic morphologies inspired by \emph{ARC Raiders}: the hexapods \emph{Queen}, \emph{Bastion}, and \emph{Tick}, and the quadruped \emph{Leaper}. The four robots span a deliberate range of leg counts, joints per leg, body sizes, and target gaits, but share a unified observation template, action convention, simulation cadence, and reward structure. We additionally provide hand-crafted Central Pattern Generator (CPG) demonstrators that serve both as fixed expert references and as sources of prior data for offline-to-online training. On this playground we conduct a controlled empirical study of how standard online RL algorithms and methods augmented with prior data behave under animation-style stylistic constraints.
}

\begin{document}

\maketitle  

\begin{abstract}
Reinforcement learning for legged locomotion has matured into a stack of multi-component reward functions and physics-engine benchmarks whose morphologies are uniformly derived from real commercial hardware. Game NPCs, however, are bound by stylistic constraints absent from sim-to-real robotics and routinely take the form of creatures with no real-robot counterpart. We introduce \textbf{\nomealgo}, a suite of four MuJoCo continuous-control environments featuring robotic morphologies inspired by the bestiary of \emph{ARC Raiders}: the 18-DoF tall hexapod \emph{Queen}, the 12-DoF armoured hexapod \emph{Bastion}, the 18-DoF compact hexapod \emph{Tick}, and the 12-DoF quadruped \emph{Leaper}. All four robots share a unified observation template, action convention, simulation cadence, and a single closed-form multi-component reward function whose only per-morphology variation lives in a small set of weights and parameters. The reward fuses a velocity-tracking tent, a healthy survive bonus, a phase-locked gait-compliance bonus/cost pair, action regularisers, three safety penalties, and a posture anchor; no motion-capture data enters the reward at any point. We additionally provide hand-crafted Central Pattern Generator demonstrators per morphology, which serve both as fixed expert references and as sources of prior data for offline-to-online training. On this playground we conduct a controlled empirical study comparing standard online algorithms (SAC, SPEQ, SOPE-EO) and methods augmented with prior data (SACfD, SPEQ-O2O, SOPE), and characterise how each paradigm copes with the playground's morphological diversity and animation-style stylistic constraints.
Source code is available at \url{https://github.com/CarloRomeo427/ARC_RL.git}.
\end{abstract}

\section{Introduction}
\label{sec:arc_rl_intro}

Since the very inception of modern deep reinforcement learning, video games have been the privileged proving ground of the field. From early arcade landmarks and perfect-information board games~\citep{mnih2015human, bellemare2013ale, silver2016mastering, schrittwieser2020mastering}, through breakthroughs in real-time strategy with AlphaStar in \emph{StarCraft~II}~\citep{vinyals2019grandmaster}, to the photorealistic driving of Gran~Turismo Sophy~\citep{wurman2022granturismo}, the history of deep RL is closely tied to game benchmarks. In parallel to RL as a player, RL has also matured as a development tool within game studios for automated playtesting and producing adaptive, learning-based behaviors for non-player characters (NPCs)~\citep{sestini2022rlgame, juliani2020unityml}.

The animation of legged characters through learned controllers has emerged as one of the most active intersections between RL and physics-based computer animation~\citep{peng2017deeploco, peng2018deepmimic, peng2021amp, peng2022ase}. By combining task and style rewards, regularizers, and phase-conditioned curricula, researchers have achieved highly agile, sim-to-real locomotion on legged platforms, and even parkour-style behaviors~\citep{hwangbo2019learning, siekmann2021periodic, cheng2024extremeparkour}. The continuous-control benchmarks that underwrite this progress---such as DM~Control~\citep{tassa2018dmcontrol}, Isaac Gym~\citep{makoviychuk2021isaacgym}, and MuJoCo~Playground~\citep{zakka2025mujocoplayground}---now constitute the natural substrate for training animation-style RL controllers.

The transposition of these techniques to commercial games, however, introduces a distinct set of constraints that the existing benchmarks address only partially. Game NPCs are subject to \emph{stylistic} requirements that have no direct counterpart in the sim-to-real robotics literature: their motion must be visually believable to a human player at every frame, must avoid the erratic, ``drunken'' gaits characteristic of bare-bones MuJoCo-Ant-style rewards~\citep{brockman2016openaigym, towers2024gymnasium}, and must conform to the aesthetic and narrative identity of the game. Moreover, commercial games routinely populate their worlds with creatures whose morphologies do not correspond to any real robot, whereas the morphologies of the canonical legged-RL benchmarks are usually derived from real commercial hardware, and therefore inherit a narrow distribution of leg counts, body plans, and proportions.

A representative example of the gap above is Embark Studios' \emph{ARC Raiders}~\citep{embark2025arcraiders}, a PvPvE extraction shooter released in October 2025 whose principal antagonists are a bestiary of legged mechanical creatures spanning scales from human-sized to mechanical giants. These adversaries simultaneously embody the two challenges identified above: they exhibit \emph{non-standard morphologies} with leg counts and proportions outside the range of any commercial robot, and they are bound by \emph{strict stylistic constraints}, because gaits that look mechanically uncanny would break player immersion in a shooter whose tension depends on the believability of its enemies. While we make no claim about the specific control techniques used in the shipped game, the design space exemplified by \emph{ARC Raiders} cleanly motivates the research question of this paper: how should existing RL paradigms be adapted---or augmented with priors---when the target is not merely a locomoting agent, but a locomoting agent with a complex morphology and a designer-specified ``look''?

To investigate this question in a reproducible setting, we introduce \textbf{\nomealgo}, a custom suite of continuous-control simulation environments featuring four robotic morphologies inspired by \emph{ARC Raiders}: the 18-DoF tall hexapod \emph{Queen}, the 12-DoF armored hexapod \emph{Bastion}, the 18-DoF compact hexapod \emph{Tick}, and the 12-DoF quadruped \emph{Leaper}. On this playground we conduct a controlled empirical study of how existing RL paradigms cope with the morphological and stylistic constraints described above, comparing standard online algorithms, online algorithms augmented with prior data, and the custom methods SPEQ~\citep{romeo2025speq} and SOPE.

\begin{figure}[t]
    \centering
    \begin{subfigure}[b]{0.24\textwidth}
        \centering
        \includegraphics[width=\textwidth]{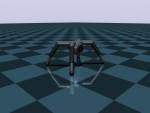}
        \caption{\textit{Leaper}}
        \label{fig:sub1}
    \end{subfigure}
    \hfill
    \begin{subfigure}[b]{0.24\textwidth}
        \centering
        \includegraphics[width=\textwidth]{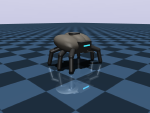}
        \caption{\textit{Bastion}}
        \label{fig:sub2}
    \end{subfigure}
    \hfill
    \begin{subfigure}[b]{0.24\textwidth}
        \centering
        \includegraphics[width=\textwidth]{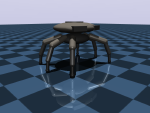}
        \caption{\textit{Queen}}
        \label{fig:sub3}
    \end{subfigure}
    \hfill
    \begin{subfigure}[b]{0.24\textwidth}
        \centering
        \includegraphics[width=\textwidth]{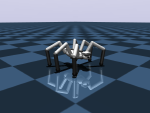}
        \caption{\textit{Tick}}
        \label{fig:sub4}
    \end{subfigure}
    \caption{\textbf{The four \nomealgo~morphologies.} Isometric renders of the robots that make up the playground: (a) \textit{Leaper}, a 12-DoF quadruped with three-link legs; (b) \textit{Bastion}, a 12-DoF armoured hexapod with two-link legs; (c) \textit{Queen}, an 18-DoF tall hexapod with three-link legs; and (d) \textit{Tick}, a compact 18-DoF hexapod sharing Queen's kinematics at a smaller scale.}
    \label{fig:four_panel}
\end{figure}

\section{Related Work}
\label{sec:arc_rl_related}

The \nomealgo~playground sits at the intersection of three lines of research: (i) the use of games and game engines as benchmarks for reinforcement learning, (ii) reward design for animation-style legged locomotion, and (iii) the use of Central Pattern Generators (CPGs) both as inductive biases inside the policy and as demonstrators for prior-data RL.

\textbf{Gaming benchmarks and physics playgrounds.}
The history of deep RL has been organised, to a remarkable degree, around the conquest of game benchmarks. Arcade games~\citep{bellemare2013ale, mnih2015human}, perfect-information board games~\citep{silver2016mastering, silver2017mastering, silver2018general, schrittwieser2020mastering}, real-time strategy titles~\citep{berner2019dota2, vinyals2019grandmaster}, and more recently procedurally generated open-world 3D environments~\citep{baker2022vpt, hafner2025dreamerv3, sima2024} have each, in turn, defined the frontier of what RL agents can do. A parallel strand has produced single-environment capability spectra in which one benchmark probes a wide range of skills~\citep{hafner2021crafter}. A more recent meaning of ``game playground'' has emerged from the use of commercial game and physics engines as embodied-AI substrates, with Unity ML-Agents~\citep{juliani2020unityml} on the game-engine side and a family of MuJoCo-based platforms~\citep{tassa2018dmcontrol, freeman2021brax, makoviychuk2021isaacgym, zakka2025mujocoplayground} on the physics side now defining the standard stack for legged-locomotion RL research. \nomealgo~inherits this stack directly---it is a MuJoCo-based benchmark in the lineage of DM~Control and MuJoCo~Playground---but departs from prior work in the origin of its morphologies: existing locomotion benchmarks bundle robots derived from real commercial hardware (ANYmal~\citep{hutter2016anymal}, Unitree Go1~\citep{unitree_go1}, Boston Dynamics Spot~\citep{bostondynamics_spot}, Cassie~\citep{xie2018cassie}, Barkour~\citep{caluwaerts2023barkour}) and therefore inherit a narrow distribution of leg counts, body plans, and proportions, whereas \nomealgo~is, to our knowledge, the first such benchmark whose morphologies are explicitly designed to mirror the agents of a contemporary commercial video game~\citep{embark2025arcraiders}.

\textbf{Reward design for legged locomotion and animation.}
Modern reinforcement learning for legged locomotion has replaced the simple reward functions of early MuJoCo Gym environments~\citep{brockman2016openaigym, towers2024gymnasium} with a standardized, multi-part reward structure. Developed through earlier sim-to-real research~\citep{hwangbo2019learning, lee2020learning, miki2022learning} and formalized by Rudin et al.~\citep{rudin2022walk}, this current standard combines terms for the task, style, regularization, safety, and survival. This decomposition produces locomotion, but not necessarily \emph{stylised} locomotion. Two complementary lines of work address this: the first, originating in computer graphics, frames stylisation as imitation of reference motion-capture clips~\citep{peng2017deeploco, peng2018deepmimic, peng2020imitating, peng2021amp, peng2022ase}; the second formalises gaits as phase-indexed contact schedules whose adherence is rewarded directly, without any reference data~\citep{siekmann2021periodic, shao2022phase, margolis2023walk}. \nomealgo~inherits the multi-component decomposition unchanged and places its animation specification in the second family: a phase-locked gait-compliance term defined against a fixed per-morphology contact schedule, with no motion-capture clip involved.

\textbf{Central Pattern Generators in legged RL.}
Central Pattern Generators (CPGs)---spinal-cord-inspired networks of coupled oscillators~\citep{ijspeert2008}---have guided legged robotics for decades. In modern deep RL, CPGs generally assume one of three roles. First, they can act as an \emph{action prior} whose intrinsic parameters are modulated by the policy, occasionally alongside residual corrections~\citep{iscen2018pmtg, bellegarda2022cpgrl, bellegarda2024visualcpgrl, shafiee2024manyquadrupeds, zhang2023synloco}. Second, they can serve as a \emph{reference trajectory} for imitation-style reward optimization~\citep{shao2022phase, li2024aicpg}, though this is less common. Finally, they can run \emph{outside} the policy as a pure \emph{phase clock} to define target stance and swing windows for reward shaping~\citep{siekmann2021periodic, siekmann2021stairs, margolis2023walk}. \nomealgo~falls squarely in the third paradigm: the CPG runs entirely outside the policy as a phase clock, and its phase variable together with per-leg offsets defines the stance and swing windows used for reward shaping.

\section{Environments}
\label{sec:arc_rl_envs}

\nomealgo~ships four MuJoCo~\citep{todorov2012mujoco} environments, exposed through the standard Gymnasium API~\citep{towers2024gymnasium}. Each environment instantiates one of four robot morphologies inspired by \emph{ARC Raiders}: \emph{Bastion}, a 12-DoF armoured hexapod with two-link legs; \emph{Queen}, an 18-DoF tall hexapod with three-link legs; \emph{Tick}, a compact 18-DoF hexapod sharing Queen's kinematics at a smaller scale; and \emph{Leaper}, a 12-DoF quadruped with three-link legs. The four robots span a deliberate range of leg counts (4 and 6), joints per leg (2 and 3), body sizes, and target gaits, but share the same observation template, the same action convention, and the same simulation cadence.

\textbf{Observation space.}
Every environment returns an identical, standardized proprioceptive observation. To ensure the learning process is stable and the state is easily interpretable, this observation is concatenated into four distinct blocks:

\begin{itemize}
    \item \textbf{Position and Posture:} Describes the robot's physical configuration. This includes the torso's height above the ground plane ($z_t$), its 3D orientation as a unit quaternion ($q^{\mathrm{base}}_t$), and the angles of all actuated joints ($q^{\mathrm{joints}}_t$). To prevent the policy from overfitting to absolute locations, horizontal world coordinates are explicitly excluded.
    \item \textbf{Velocities:} Tracks the robot's momentum for dynamic balancing. This consists of the torso's linear ($v^{\mathrm{base}}_t$) and angular ($\omega^{\mathrm{base}}_t$) velocities, alongside the rotational speeds of each joint ($\dot{q}^{\mathrm{joints}}_t$).
    \item \textbf{Contact Forces:} Monitors physical interactions with the ground. It measures the external forces and torques acting on each of the robot's rigid bodies ($c^{\mathrm{ext}}_t$). To protect the value function from extreme mathematical spikes during hard impacts, these variables are strictly clipped to the interval $[-1, 1]$.
    \item \textbf{Gait Phase Clock:} Informs the robot of its current step cycle. A continuous timing variable ($\phi_t \in [0, 2\pi)$) dictates the rhythm of the gait. To ensure the neural network interprets this smoothly without a mathematical discontinuity when the clock resets, it is encoded as $[\sin\phi_t, \cos\phi_t]$.
\end{itemize}

Depending on the specific robot's joint and body counts, this structured formulation yields an observation dimensionality of $151$ for Bastion, $163$ for Leaper, and $199$ for Queen and Tick.

\textbf{Action space.}
Each robot is driven by direct joint-level torque-equivalent commands. Actions live in $\mathcal{A} = [-1, 1]^{n_u}$, where $n_u$ is the number of actuated joints---$12$ for Bastion and Leaper, $18$ for Queen and Tick---and each scalar is multiplied inside MuJoCo by a per-actuator \texttt{gear} constant to produce the applied torque.

\textbf{Simulation step and termination.}
All four environments run at the same simulation frequency, with a fixed frame-skip of $25$ MuJoCo substeps per control step. An episode terminates early when the torso $z$-coordinate leaves a per-robot healthy range, which is the only failure condition; otherwise, episodes run to a fixed time horizon. Each environment exposes a small set of per-morphology parameters---target velocity, gait frequency, healthy $z$-range, foot stance threshold---that the reward function consumes.

\section{Reward Function}
\label{sec:arc_rl_reward}

The reward function is structurally identical across the four morphologies: a single closed-form expression, one implementation, one set of named terms, with per-robot adaptation expressed only through weights and parameters. At every step the policy receives:
\begin{equation}
\label{eq:arc_rl_reward}
r_t \;=\; r_{\mathrm{fwd}} \;+\; r_{\mathrm{h}} \;+\; r_{\mathrm{gait}}^{+} \;-\; \big( c_{\mathrm{gait}} + c_{\mathrm{ctrl}} + c_{\mathrm{smooth}} + c_{\mathrm{contact}} + c_{\mathrm{ang}} + c_{\mathrm{zvel}} + c_{\mathrm{post}} \big).
\end{equation}
The three positive terms encode \emph{what we want}: $r_{\mathrm{fwd}}$ rewards moving at a target speed, $r_{\mathrm{h}}$ rewards staying upright, and $r_{\mathrm{gait}}^{+}$ rewards matching a prescribed contact pattern. The seven negative terms encode \emph{what we want to avoid}: $c_{\mathrm{gait}}$ penalises deviations from the prescribed contact pattern; the remaining six are regularisers and safety penalties. We describe all ten terms in turn.

\textbf{Forward velocity tracking.}
The task term shapes the speed of the desired animation. It is a symmetric triangular tent over the forward body velocity $v_x$, peaked at a per-robot target speed $v^\star$ and identically zero outside a fixed band $[v^\star - \sigma_v, v^\star + \sigma_v]$:
\begin{equation}
r_{\mathrm{fwd}}(v_x) \;=\; w_{\mathrm{fwd}} \cdot \max\!\Big(0,\; 1 - |v_x - v^\star| / \sigma_v\Big).
\end{equation}
The standard choice in the modern legged-RL recipe is a Gaussian bell~\citep{rudin2022walk}; \nomealgo~substitutes a triangular tent for two reasons. First, the tent has \emph{compact support}: it is exactly zero outside the band $[v^\star - \sigma_v, v^\star + \sigma_v]$, whereas the Gaussian has long tails that contribute small but non-zero positive reward to any velocity. This gives the per-step task reward a finite, well-defined maximum and a clean composition with the other reward terms. Second, the compact support keeps the velocity term from competing with the gait, posture, and safety terms outside the intended speed band---exactly the regime in which those terms carry most of the gradient signal.

\textbf{Healthy survive bonus.}
The safety term is a constant bonus paid at every step in which the torso $z$-coordinate lies in the per-robot healthy range $[z_{\min}, z_{\max}]$:
\begin{equation}
r_{\mathrm{h}} \;=\; w_{\mathrm{h}} \cdot \mathbf{1}\!\big[\,z_{\mathrm{torso}} \in [z_{\min}, z_{\max}]\,\big].
\end{equation}
When the condition is violated the episode terminates and no further reward is collected, so $r_{\mathrm{h}}$ acts both as a survive bonus and as the trigger for the Early Termination mechanism of DeepMimic~\citep{peng2018deepmimic}.

\textbf{Gait compliance.}
The style term shapes the \emph{contact pattern} of the desired animation: which feet are supposed to be touching the ground at any given moment. We refer to a foot as being ``in stance'' when it is on the ground and ``in swing'' when it is in the air; the pattern of stance and swing across all feet over a gait cycle is what makes a gait look like an alternating tripod, a trot, or anything else.

To define stance and swing, we introduce three elements that the reward needs but that do not otherwise appear in the observation. The first is a phase clock $\phi \in [0, 2\pi)$ that advances at a fixed gait frequency $f_g$ and tells the reward function ``where in the gait cycle we currently are''. The second is a duty fraction $d \in (0, 1)$ that splits each cycle into a stance portion ($\phi$ small) and a swing portion ($\phi$ large); $d = 0.6$ means each foot is supposed to spend 60\% of its cycle on the ground. The third is a per-foot phase offset $\Delta_i$ that staggers the feet across the cycle: the alternating tripod, for example, puts half the legs in stance while the other half is in swing.

Given these three elements, the \emph{target} stance state of foot $i$ at phase $\phi$ is
\begin{equation}
\mathrm{stance}_i^{\mathrm{tgt}}(\phi) \;=\; \mathbf{1}\!\big[\,(\phi + \Delta_i) \bmod 2\pi < 2\pi d\,\big],
\end{equation}
and the \emph{actual} stance state is obtained by thresholding the foot height with a per-robot threshold $z_{\mathrm{thr}}$,
\begin{equation}
\mathrm{stance}_i^{\mathrm{act}} \;=\; \mathbf{1}\!\big[\,z_{\mathrm{foot},i} < z_{\mathrm{thr}}\,\big].
\end{equation}
The reward then counts the number of feet whose actual state disagrees with the target,
\begin{equation}
N_e(\phi) \;=\; \sum_{i=1}^{N} \mathbf{1}\!\big[\,\mathrm{stance}_i^{\mathrm{act}} \ne \mathrm{stance}_i^{\mathrm{tgt}}(\phi)\,\big],
\end{equation}
and uses this single quantity to construct two complementary terms:
\begin{equation}
r_{\mathrm{gait}}^{+}(\phi) \;=\; w_{\mathrm{gb}} \cdot \Big(1 - \tfrac{N_e}{N}\Big),
\qquad
c_{\mathrm{gait}}(\phi) \;=\; w_{\mathrm{gc}} \cdot N_e.
\end{equation}
The bonus $r_{\mathrm{gait}}^{+}$ is normalised by the number of feet $N$, so it always lives in $[0, w_{\mathrm{gb}}]$ regardless of morphology and is maximised when every foot is in the correct mode at the current phase. The cost $c_{\mathrm{gait}}$ is un-normalised and grows linearly with the number of mismatched feet. We use a bonus--cost pair rather than a single signed term for three reasons: the bounded bonus places the gait reward on the same scale as the task term across morphologies; the un-normalised cost makes large persistent errors progressively more expensive; and the pair structure doubles the gradient signal at each foot's stance/swing boundary compared with a single signed term. Together they implement a deterministic-indicator analogue of the periodic reward composition of Siekmann et al.~\citep{siekmann2021periodic}. The choice of contact pattern is encoded entirely in the per-foot offsets $\{\Delta_i\}$: an alternating tripod for the three hexapods, a diagonal-pair trot for Leaper.

\textbf{Action regularisers.}
Two regularisers shape the control signal: a magnitude penalty and a smoothness penalty,
\begin{equation}
c_{\mathrm{ctrl}} \;=\; w_{\mathrm{c}} \cdot \lVert a_t \rVert_2^2,
\qquad
c_{\mathrm{smooth}} \;=\; w_{\mathrm{s}} \cdot \lVert a_t - a_{t-1} \rVert_2^2.
\end{equation}
The first is the L2 action penalty inherited from the Gymnasium-Ant template~\citep{brockman2016openaigym, towers2024gymnasium} and promotes minimum-energy solutions; the second is the temporal smoothness regulariser of Mysore et al.~\citep{mysore2021caps} and prevents high-frequency, jittery control signals. The previous action $a_{t-1}$ is maintained internally by the environment and is not exposed to the policy.

\textbf{Safety penalties.}
Three further costs discourage motions that are unsafe or visually wrong:
\begin{equation}
c_{\mathrm{contact}} \;=\; w_{\mathrm{cc}} \cdot \big\lVert \mathrm{clip}\!\big(c^{\mathrm{ext}}_t,\, [-1,1]\big) \big\rVert_2^2,
\qquad
c_{\mathrm{ang}} \;=\; w_{\mathrm{a}} \cdot \big(\omega_{\mathrm{roll}}^2 + \omega_{\mathrm{pitch}}^2\big),
\qquad
c_{\mathrm{zvel}} \;=\; w_{\mathrm{z}} \cdot (v^{\mathrm{base}}_{z,t})^2.
\end{equation}
The contact-force penalty $c_{\mathrm{contact}}$ discourages large external contact forces; the angular-velocity penalty $c_{\mathrm{ang}}$ discourages excessive body-frame roll and pitch rates (yaw is left free so the policy can steer); and the vertical-velocity penalty $c_{\mathrm{zvel}}$ discourages bouncing of the torso. The three follow the safety template of Rudin et al.~\citep{rudin2022walk} and target failure modes (high-impact landings, body twisting, bounding) that are otherwise compatible with positive forward and gait reward.

\textbf{Posture anchor.}
The final cost is a quadratic anchor toward a hand-designed default joint configuration $q^{\mathrm{def}}$,
\begin{equation}
c_{\mathrm{post}} \;=\; w_{\mathrm{p}} \cdot \lVert q^{\mathrm{joints}}_t - q^{\mathrm{def}} \rVert_2^2.
\end{equation}
Without this term, policies routinely discover configurations (splayed legs, inverted knees) that satisfy the gait reward but produce visually wrong animations. The posture cost follows the joint nominal pose term of Hwangbo et al.~\citep{hwangbo2019learning} and Rudin et al.~\citep{rudin2022walk}, and plays a role analogous to DeepMimic's joint-pose imitation term~\citep{peng2018deepmimic} but with a constant target rather than a phase-indexed reference clip---no motion-capture data enters the reward at any point.

\textbf{Per-morphology parameters.}
All four robots use the same reward expression and the same default weights for the principal terms; per-morphology adaptation lives entirely in the target velocity $v^\star$, the velocity band width $\sigma_v$, the gait frequency $f_g$, the foot phase offsets $\{\Delta_i\}$, the foot stance threshold $z_{\mathrm{thr}}$, the healthy $z$-range $[z_{\min}, z_{\max}]$, and the control-cost weight $w_{\mathrm{c}}$ (which is scaled inversely with the action dimensionality).

\section{CPG Controllers as Expert Policies}
\label{sec:arc_rl_cpg}

For each robot, we provide an open-loop controller capable of producing a competent walking gait. Acting as an expert demonstrator, its purpose is to be a strong baseline for further comparisons, but also to populate the prior-data buffers used by our RL algorithm with baseline walking trajectories.

\textbf{The Rhythmic Engine.}
At the heart of the controller is the \emph{gait clock}---an internal metronome ticking forward at a fixed frequency. Each leg tracks this master clock but applies its own specific delay, or \emph{offset}. This staggered timing dictates the robot's overall gait pattern: applying an offset of half a cycle to alternating legs creates an alternating tripod gait for the hexapods (Bastion, Queen, Tick), or a diagonal trot for the quadruped (Leaper). The controller also uses a \emph{duty factor} to divide each leg's cycle into a stance phase (pushing on the ground) and a swing phase (lifting through the air).

\textbf{Generating Movement.}
Based on whether a leg is in stance or swing, the controller generates specific target angles for the hip, knee, and ankle joints. During stance, the hip sweeps backward to propel the body forward, while the knee and ankle provide a smooth push-off. During swing, the hip resets forward, and the lower joints lift the foot clear of the ground. These target trajectories are shaped by simple kinematic primitives---triangle waves for the hips and half-sine bells for the knees and ankles---ensuring smooth transitions when the foot strikes or leaves the ground.

\textbf{Execution.}
To translate these target angles into physical motion, the controller employs a standard proportional--derivative (PD) tracking loop. This system continuously calculates the necessary joint torques to minimize the error between the robot's current pose and the generated target angles. The resulting torques are smoothly ramped up at the start of an episode to prevent sudden jerking motions, clipped to the simulation's expected action range, and passed directly to MuJoCo. Ultimately, these hand-crafted routines provide a robust foundation of prior data to accelerate the subsequent RL training phases.

\section{Experimental Results}
\label{sec:arc_rl_experiments}

In this section we describe the experimental protocol used to evaluate the \nomealgo~playground, list the algorithms compared, and report the resulting learning curves on the four robots. The goal of this evaluation is not to crown a winning algorithm, but to characterise how the methods discussed throughout this paper---and the standard online baselines---behave on a common, animation-style legged-locomotion playground. To facilitate future research, we have released the SPEQ code-base here: \url{https://github.com/CarloRomeo427/ARC_RL.git}.

\textbf{Experimental Setting.}
We evaluate seven configurations grouped into three families. The first family contains a single \emph{expert} baseline: the CPG controllers introduced in the previous section, evaluated under the same protocol as the learned policies. The CPG is not trained, but it is plotted as a constant reference line that shows the return achievable by an open-loop, gait-correct demonstrator on each robot. The second family is the \emph{online} family, in which the agents learn purely from environment interaction without any prior data: SAC~\citep{haarnoja2018soft}, the standard sample-efficient off-policy baseline; SPEQ~\citep{romeo2025speq}, the periodic offline-stabilisation algorithm; and SOPE-EO, the online-only variant of SOPE~\citep{SOPE} in which the actor-aligned early-stopping rule to control the length of each stabilization phase is calculated exclusively on the online replay buffer. The third family is the \emph{online with prior data} family, in which the same three algorithmic ideas are combined with the CPG-generated prior buffer: SACfD~\citep{sacfd1}, the SAC-from-demonstrations baseline; SPEQ-O2O, the prior-data variant of SPEQ; and SOPE, the full algorithm.

\begin{figure}[t]
\centering
\includegraphics[width=0.6\textwidth]{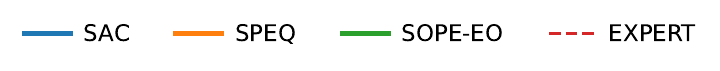}\\[0.4em]
\begin{minipage}[t]{0.24\textwidth}
    \centering
    {\small\textbf{Leaper}}\\[2pt]
    \includegraphics[width=\linewidth]{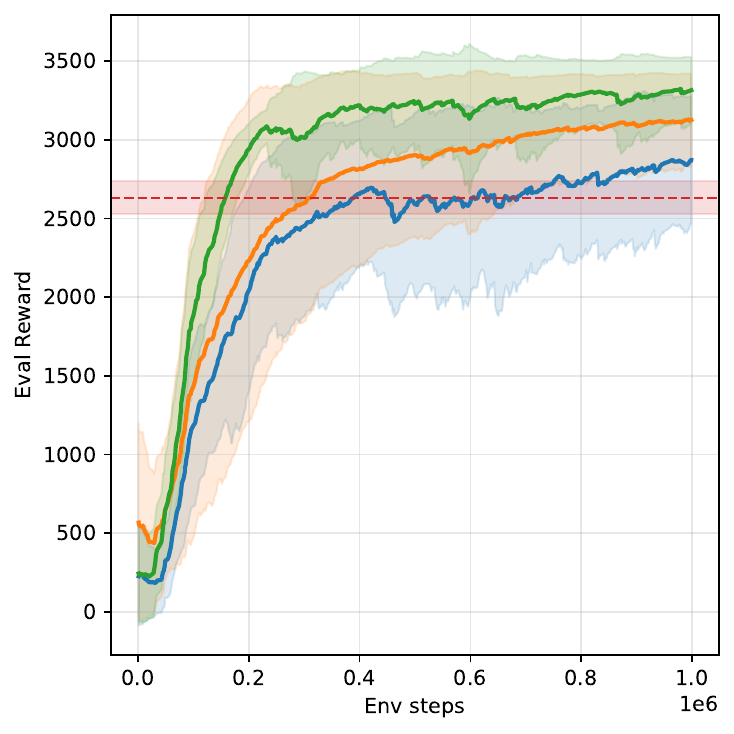}
\end{minipage}\hfill
\begin{minipage}[t]{0.24\textwidth}
    \centering
    {\small\textbf{Bastion}}\\[2pt]
    \includegraphics[width=\linewidth]{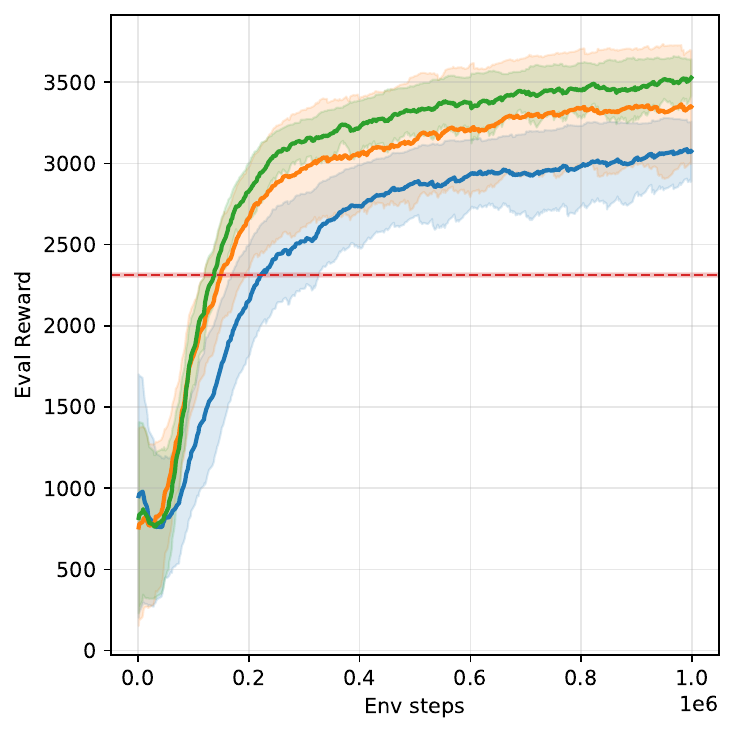}
\end{minipage}\hfill
\begin{minipage}[t]{0.24\textwidth}
    \centering
    {\small\textbf{Queen}}\\[2pt]
    \includegraphics[width=\linewidth]{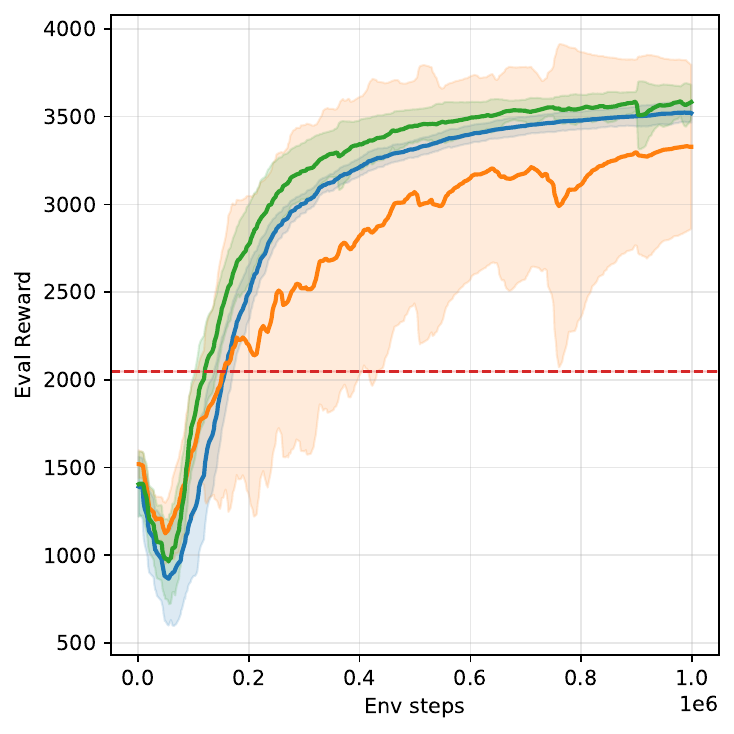}
\end{minipage}\hfill
\begin{minipage}[t]{0.24\textwidth}
    \centering
    {\small\textbf{Tick}}\\[2pt]
    \includegraphics[width=\linewidth]{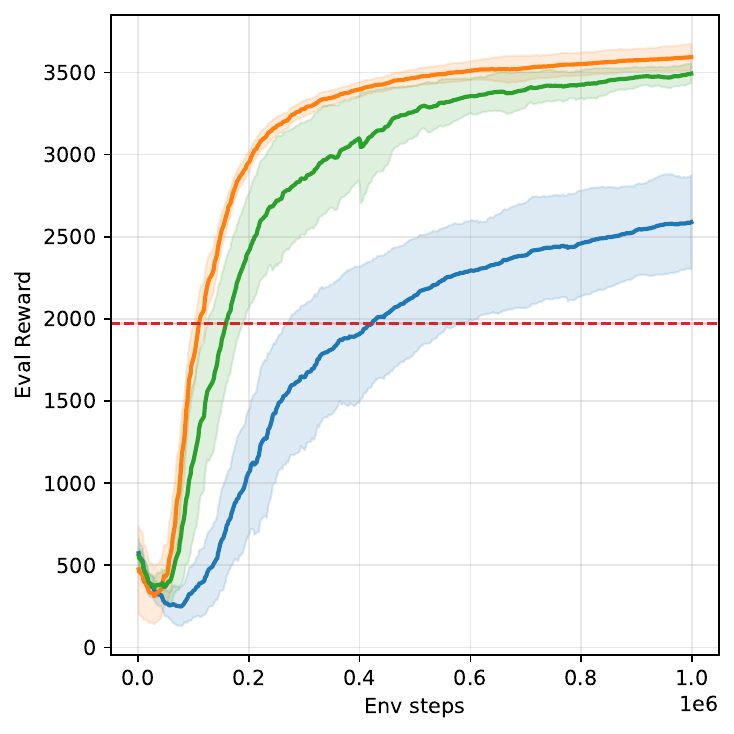}
\end{minipage}
\caption{\textbf{Online RL on the four \nomealgo~robots.} Evaluation returns as a function of environment steps for SAC, SPEQ, and SOPE-EO, with the CPG controller plotted as a constant expert reference. Solid lines denote the mean across $5$ random seeds, shaded regions the standard deviation.}
\label{fig:arc_rl_online}
\end{figure}

\begin{figure}[t]
\centering
\includegraphics[width=0.6\textwidth]{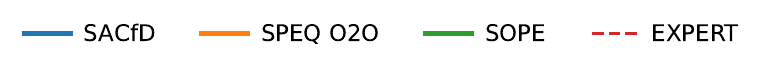}\\[0.4em]
\begin{minipage}[t]{0.24\textwidth}
    \centering
    {\small\textbf{Leaper}}\\[2pt]
    \includegraphics[width=\linewidth]{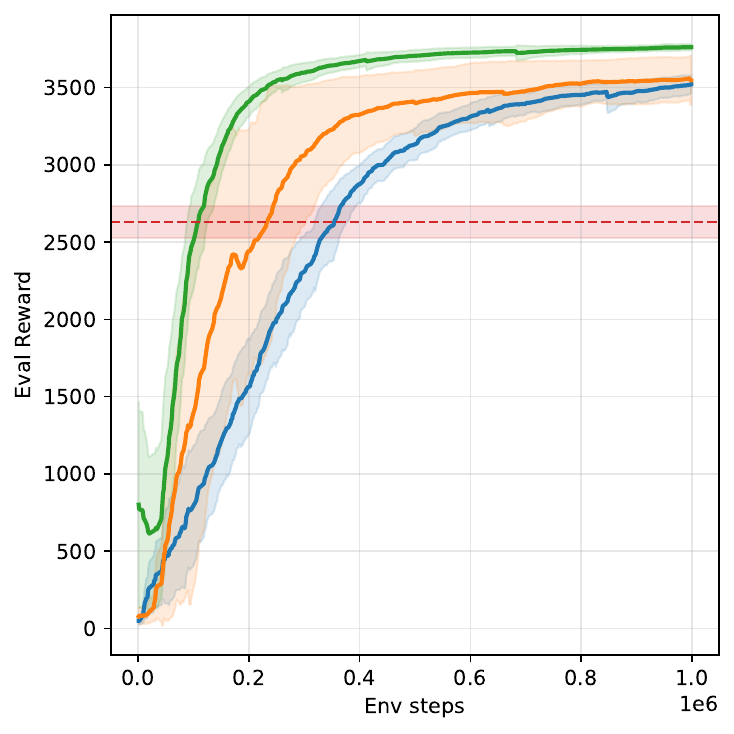}
\end{minipage}\hfill
\begin{minipage}[t]{0.24\textwidth}
    \centering
    {\small\textbf{Bastion}}\\[2pt]
    \includegraphics[width=\linewidth]{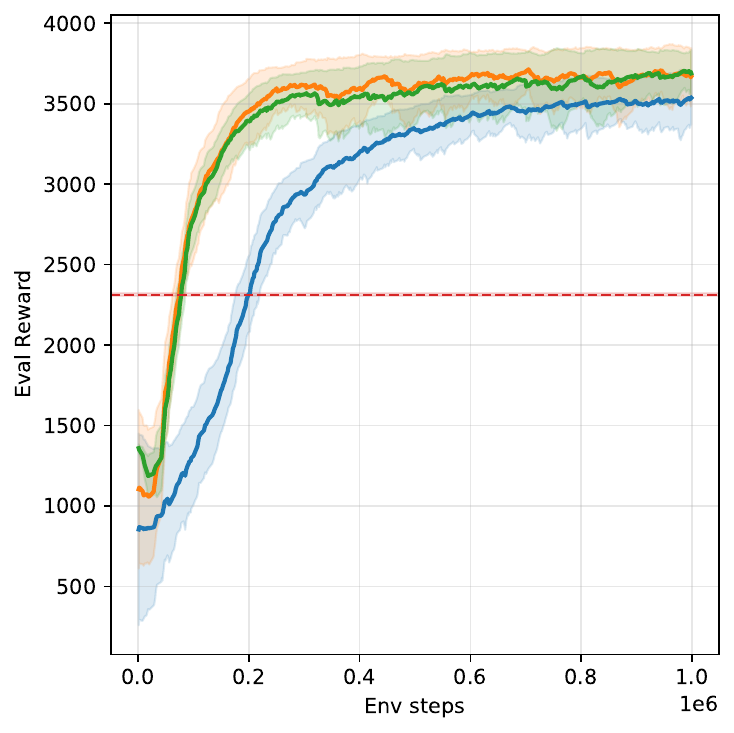}
\end{minipage}\hfill
\begin{minipage}[t]{0.24\textwidth}
    \centering
    {\small\textbf{Queen}}\\[2pt]
    \includegraphics[width=\linewidth]{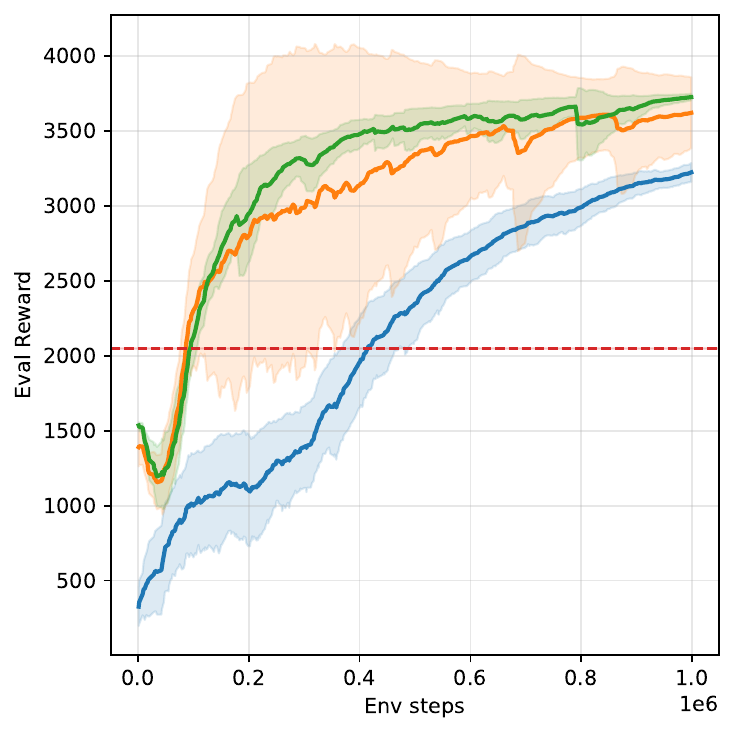}
\end{minipage}\hfill
\begin{minipage}[t]{0.24\textwidth}
    \centering
    {\small\textbf{Tick}}\\[2pt]
    \includegraphics[width=\linewidth]{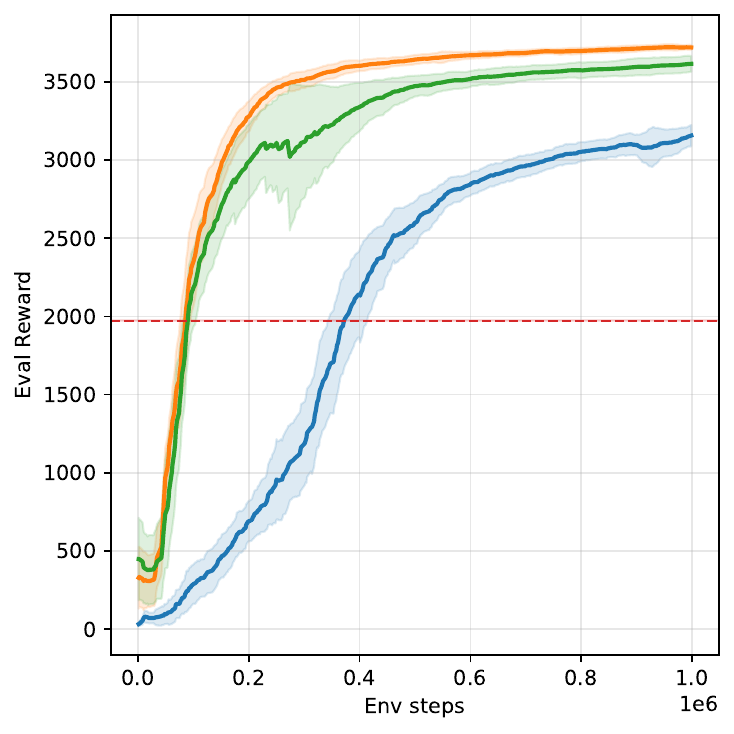}
\end{minipage}
\caption{\textbf{Online RL with prior data on the four \nomealgo~robots.} Evaluation returns as a function of environment steps for SACfD, SPEQ-O2O, and SOPE, each consuming the CPG-generated prior buffer, with the CPG controller plotted as a constant expert reference. Solid lines denote the mean across $5$ random seeds, shaded regions the standard deviation.}
\label{fig:arc_rl_online_prior}
\end{figure}

All learned policies are trained for $1$ million environment steps. Episodes are capped at the per-robot time horizon of Section~\ref{sec:arc_rl_envs} and terminate early when the torso $z$-coordinate leaves the healthy range. We evaluate every $10{,}000$ environment steps by rolling out the current policy for $10$ episodes with deterministic actions and reporting the mean episodic return. Every configuration is run with $5$ random seeds; the curves shown in Figures~\ref{fig:arc_rl_online} and~\ref{fig:arc_rl_online_prior} report the mean across seeds together with the standard deviation as a shaded band. The CPG reference line is the average return of the controller across $5$ rollouts, also with a shaded standard-deviation band.

\textbf{Online RL.}
Figure~\ref{fig:arc_rl_online} reports the learning curves for the online family on the four robots. Within the one-million-step budget all three algorithms surpass the CPG reference line on every morphology, indicating that the playground is solvable by pure online RL and that the learned policies, given enough interaction, exceed the open-loop demonstrator built into the reward function. In addition, SOPE-EO performs the best across almost all the environments, surpassed by very few points by SPEQ only in the Tick environment. On the other hand, SAC's performance is competitive only in the Queen's environment where it matches SOPE-EO.

\textbf{Online RL with prior data.}
Figure~\ref{fig:arc_rl_online_prior} reports the corresponding curves for the prior-data family.
The prior data generated via the CPG controllers contributes in two clear ways.
First, the three algorithms start from a noticeably higher initial return on every robot than their online-only counterparts, reflecting the fact that providing prior knowledge is strongly beneficial for the sample efficiency of off-policy algorithms. Second, the final scores of all three algorithms are higher than those of the online counterpart, highlighting once again the benefits of leveraging a prior distribution.

\begin{figure}[t]
     \centering
     \begin{subfigure}[b]{0.3\textwidth}
         \centering
         \includegraphics[width=\textwidth]{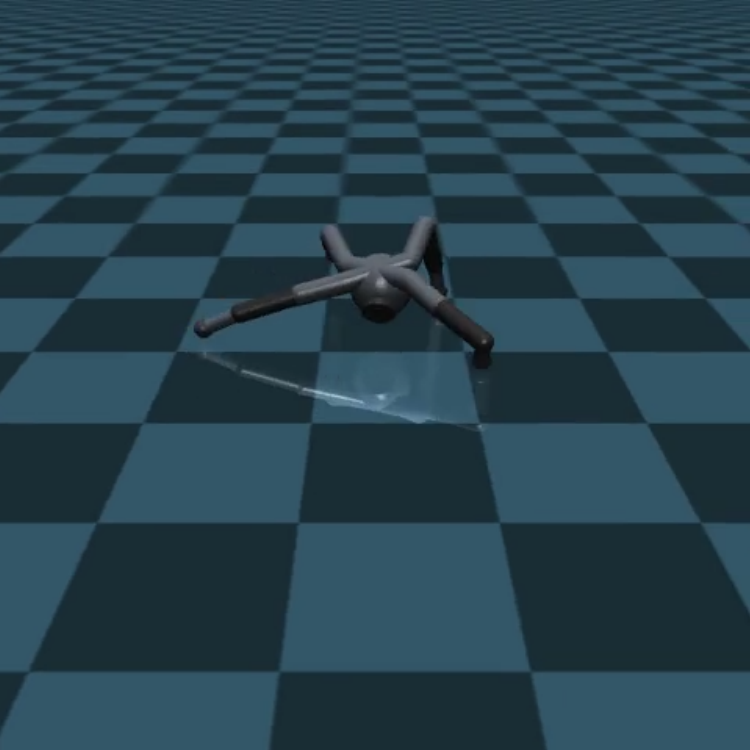}
         \caption{SAC}
     \end{subfigure}
     \hfill
     \begin{subfigure}[b]{0.3\textwidth}
         \centering
         \includegraphics[width=\textwidth]{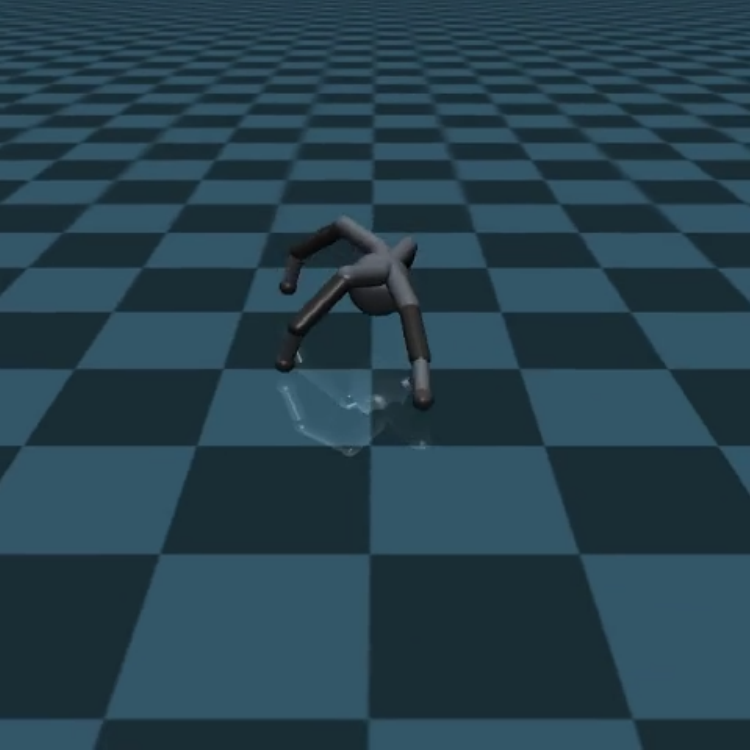}
         \caption{SPEQ}
     \end{subfigure}
     \hfill
     \begin{subfigure}[b]{0.3\textwidth}
         \centering
         \includegraphics[width=\textwidth]{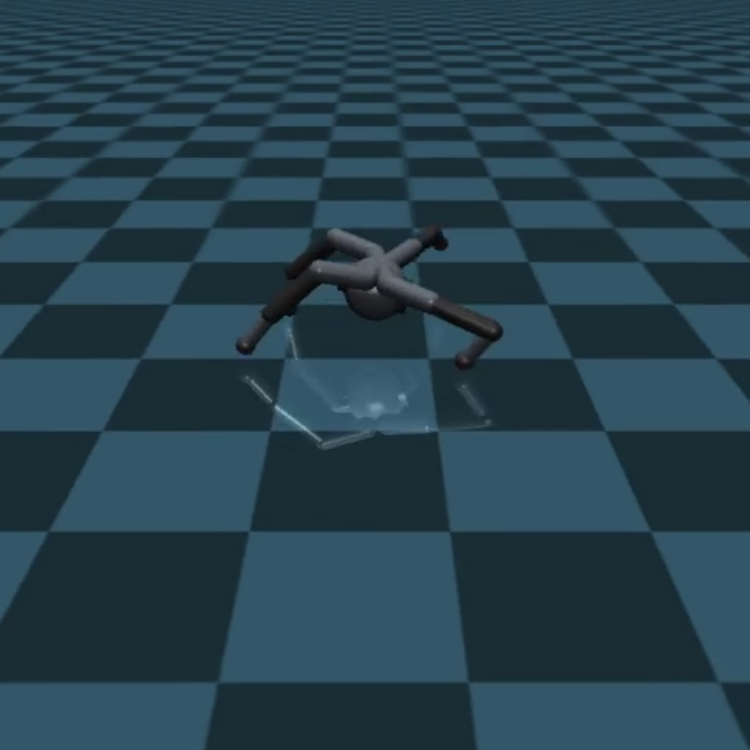}
         \caption{SOPE EO}
     \end{subfigure}

     \vspace{0.5cm} 

     \begin{subfigure}[b]{0.3\textwidth}
         \centering
         \includegraphics[width=\textwidth]{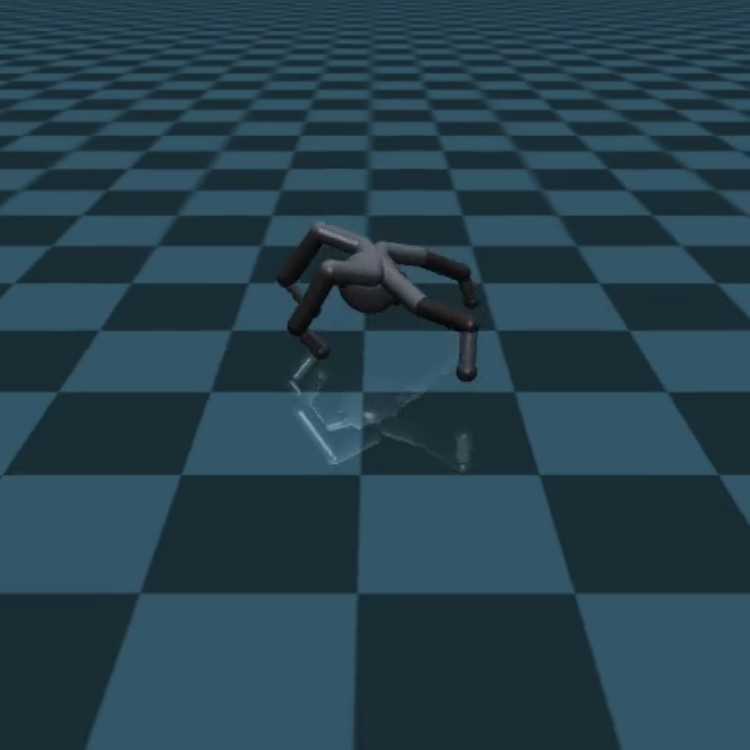}
         \caption{SACfD}
     \end{subfigure}
     \hfill
     \begin{subfigure}[b]{0.3\textwidth}
         \centering
         \includegraphics[width=\textwidth]{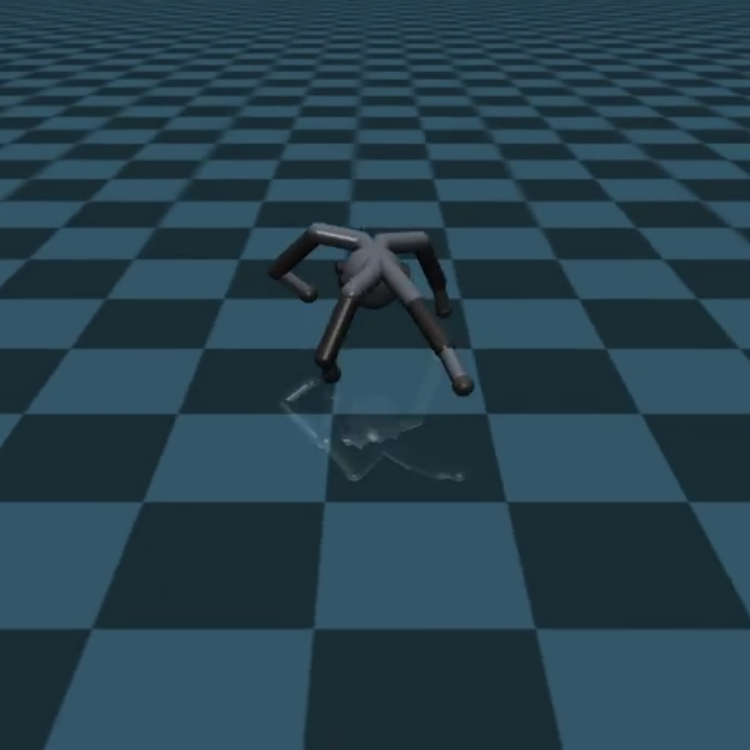}
         \caption{SPEQ O2O}
     \end{subfigure}
     \hfill
     \begin{subfigure}[b]{0.3\textwidth}
         \centering
         \includegraphics[width=\textwidth]{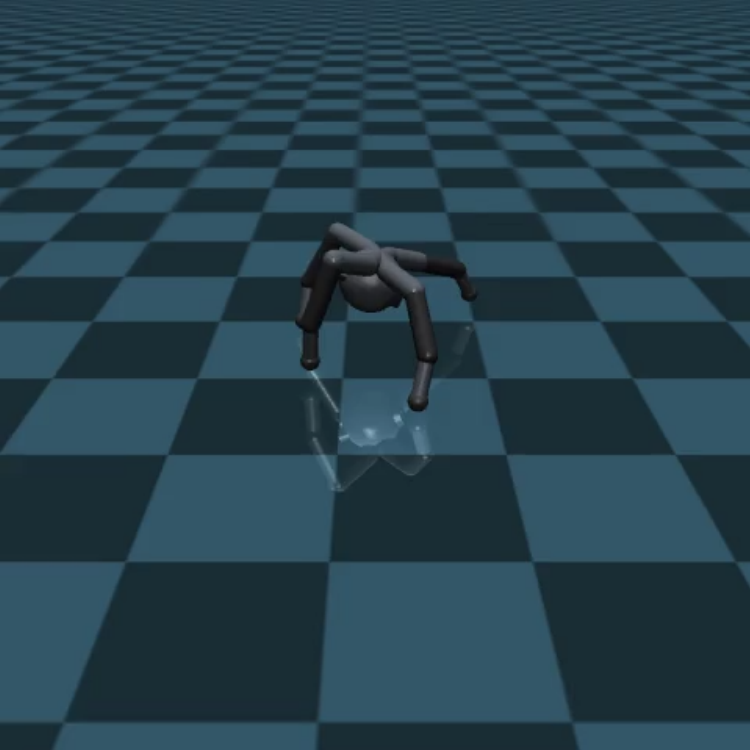}
         \caption{SOPE}
     \end{subfigure}

     \caption{\textbf{Visual comparison of \textit{Leaper} policies across algorithms.}
    Representative frames captured at equivalent points of the gait cycle during evaluation. The top row shows the exclusively online algorithms (SAC, SPEQ, SOPE-EO); the bottom row shows their counterparts augmented with prior data (SACfD, SPEQ-O2O, SOPE). The round black ``eye'' on the front of the chassis indicates the intended forward-facing direction, which serves as a visual reference for heading alignment. Prior-data methods (bottom row) reproduce the diagonal-trot gait and stylistic constraints encoded by the reward function more faithfully than their online-only counterparts, with SOPE achieving the closest match to the reference animation. Frames are illustrative of a single sequence and should not be generalised to all evaluation rollouts.}
    \label{fig:ARC_comparison}
\end{figure}

\textbf{A visual comparison of the generated animations.}
While evaluating animation quality from static frames is inherently limited, visual comparisons at equivalent points in the gait cycle reveal distinct behavioral patterns. Figure~\ref{fig:ARC_comparison} illustrates representative frames captured during the evaluation phase of the \textit{Leaper} environment. Although these observations are specific to this sequence and should not be generalized across all experiments without further analysis, they provide clear intuition into how each algorithm shapes the agent's posture and locomotion.
The following analysis aims to provide a more comprehensive picture than the previous comparison of learning curves alone.
The figure is divided into two rows based on the training methodology: the top row displays the exclusively online algorithms (SAC, SPEQ, and SOPE-EO), while the bottom row shows the algorithms augmented with prior data (SACfD, SPEQ-O2O, and SOPE).

\textbf{Online Solutions.}
The pure online methods deviate the furthest from the reference animation:
\begin{itemize}
    \item \textbf{SAC:} The policy fails to maintain the correct forward heading, completely rotating the model to move laterally (the front of the \textit{Leaper} is indicated by the round black ``eye'').
    \item \textbf{SPEQ:} While better aligned with the correct forward motion, the agent exhibits noticeable postural anomalies, particularly an excessive forward pitch of the torso.
    \item \textbf{SOPE-EO:} This method is the closest to the reference among the online solutions. However, it still displays an unnatural overreaching behavior, where the front legs extend too far forward to strike the ground.
\end{itemize}

\textbf{Online with Prior Data Solutions.}
The inclusion of prior data significantly improves stylistic compliance, though minor artifacts remain in the baseline methods:
\begin{itemize}
    \item \textbf{SACfD:} This robustly tracks the overall reference trajectory, but mechanical flaws persist. The rear legs remain excessively stiff, pushing the model forward rigidly, while the front legs bend unnaturally inward.
    \item \textbf{SPEQ-O2O:} The prior distribution successfully corrects the severe forward-bending issue seen in its online counterpart, though the overall stance still leaves some room for refinement.
    \item \textbf{SOPE:} This algorithm achieves the highest visual fidelity. It almost perfectly replicates the reference animation, demonstrating an effective and natural use of the diagonal trot encoded by the reward function's gait system.
\end{itemize}

\section{Conclusion}
\label{sec:arc_rl_conclusion}

In this paper, we introduced \textbf{\nomealgo}, a continuous-control simulation playground designed to bridge the gap between standard legged-locomotion benchmarks and the specific morphological and stylistic demands of modern commercial video games. Inspired by the mechanical adversaries of \textit{ARC Raiders}, we developed four distinct robotic environments that span a deliberate range of leg counts, joints per leg, and body sizes: the hexapods Queen, Bastion, and Tick, alongside the quadruped Leaper. Alongside these environments, we provided Central Pattern Generator (CPG) demonstrators to serve as fixed expert references, generating offline datasets of experiences collected by executing these expert policies.

Furthermore, we conducted a comprehensive evaluation comparing pure online RL algorithms against methods augmented with prior data. This analysis aimed to determine how closely each paradigm approaches the expert baseline and how faithfully it reproduces the stylized animations induced by our multi-objective reward function.

Despite these contributions, the \nomealgo~project leaves room for future enhancements to match the comprehensiveness of state-of-the-art benchmark suites in the RL literature. Currently, the training environments consist exclusively of flat ground planes. Introducing surface discontinuities and procedurally generated terrain has been proven to robustly induce generalization during training and would significantly elevate the benchmark. Additionally, the suite implements only forward-locomotion tasks. Incorporating complex navigation objectives, such as steering or obstacle avoidance, would provide deeper insights into the versatile animation capabilities of each agent. In terms of prior data, the provided datasets are currently limited to CPG trajectories; incorporating rich, agent-collected datasets, similar to those found in the Minari suite, would be highly beneficial for advancing offline experimentation. Finally, while the multi-aspect reward function successfully enforces gait compliance, it can be further refined to provide a denser, more robust learning signal for the agents across a wider variety of behaviors.



\bibliography{main}

@article{mnih2015human,
  title   = {Human-level control through deep reinforcement learning},
  author  = {Mnih, Volodymyr and Kavukcuoglu, Koray and Silver, David and Rusu, Andrei A. and Veness, Joel and Bellemare, Marc G. and Graves, Alex and Riedmiller, Martin and Fidjeland, Andreas K. and Ostrovski, Georg and others},
  journal = {Nature},
  volume  = {518},
  number  = {7540},
  pages   = {529--533},
  year    = {2015},
  doi     = {10.1038/nature14236}
}

@article{bellemare2013ale,
  title   = {The Arcade Learning Environment: An Evaluation Platform for General Agents},
  author  = {Bellemare, Marc G. and Naddaf, Yavar and Veness, Joel and Bowling, Michael},
  journal = {Journal of Artificial Intelligence Research},
  volume  = {47},
  pages   = {253--279},
  year    = {2013},
  doi     = {10.1613/jair.3912}
}

@article{silver2016mastering,
  title   = {Mastering the game of {Go} with deep neural networks and tree search},
  author  = {Silver, David and Huang, Aja and Maddison, Chris J. and Guez, Arthur and Sifre, Laurent and van den Driessche, George and Schrittwieser, Julian and Antonoglou, Ioannis and Panneershelvam, Veda and Lanctot, Marc and others},
  journal = {Nature},
  volume  = {529},
  number  = {7587},
  pages   = {484--489},
  year    = {2016},
  doi     = {10.1038/nature16961}
}

@article{silver2017mastering,
  title   = {Mastering the game of {Go} without human knowledge},
  author  = {Silver, David and Schrittwieser, Julian and Simonyan, Karen and Antonoglou, Ioannis and Huang, Aja and Guez, Arthur and Hubert, Thomas and Baker, Lucas and Lai, Matthew and Bolton, Adrian and others},
  journal = {Nature},
  volume  = {550},
  number  = {7676},
  pages   = {354--359},
  year    = {2017},
  doi     = {10.1038/nature24270}
}

@article{silver2018general,
  title   = {A general reinforcement learning algorithm that masters chess, shogi, and {Go} through self-play},
  author  = {Silver, David and Hubert, Thomas and Schrittwieser, Julian and Antonoglou, Ioannis and Lai, Matthew and Guez, Arthur and Lanctot, Marc and Sifre, Laurent and Kumaran, Dharshan and Graepel, Thore and others},
  journal = {Science},
  volume  = {362},
  number  = {6419},
  pages   = {1140--1144},
  year    = {2018},
  doi     = {10.1126/science.aar6404}
}

@article{schrittwieser2020mastering,
  title   = {Mastering {Atari}, {Go}, chess and shogi by planning with a learned model},
  author  = {Schrittwieser, Julian and Antonoglou, Ioannis and Hubert, Thomas and Simonyan, Karen and Sifre, Laurent and Schmitt, Simon and Guez, Arthur and Lockhart, Edward and Hassabis, Demis and Graepel, Thore and Lillicrap, Timothy and Silver, David},
  journal = {Nature},
  volume  = {588},
  number  = {7839},
  pages   = {604--609},
  year    = {2020},
  doi     = {10.1038/s41586-020-03051-4}
}

@article{vinyals2019grandmaster,
  title   = {Grandmaster level in {StarCraft~II} using multi-agent reinforcement learning},
  author  = {Vinyals, Oriol and Babuschkin, Igor and Czarnecki, Wojciech M. and Mathieu, Micha{\"e}l and Dudzik, Andrew and Chung, Junyoung and Choi, David H. and Powell, Richard and Ewalds, Timo and Georgiev, Petko and others},
  journal = {Nature},
  volume  = {575},
  number  = {7782},
  pages   = {350--354},
  year    = {2019},
  doi     = {10.1038/s41586-019-1724-z}
}

@article{wurman2022granturismo,
  title   = {Outracing champion {Gran Turismo} drivers with deep reinforcement learning},
  author  = {Wurman, Peter R. and Barrett, Samuel and Kawamoto, Kenta and MacGlashan, James and Subramanian, Kaushik and Walsh, Thomas J. and Capobianco, Roberto and Devlic, Alisa and Eckert, Franziska and Fuchs, Florian and others},
  journal = {Nature},
  volume  = {602},
  number  = {7896},
  pages   = {223--228},
  year    = {2022},
  doi     = {10.1038/s41586-021-04357-7}
}

@article{berner2019dota2,
  title   = {Dota 2 with Large Scale Deep Reinforcement Learning},
  author  = {Berner, Christopher and Brockman, Greg and Chan, Brooke and Cheung, Vicki and D{\k{e}}biak, Przemys{\l}aw and Dennison, Christy and Farhi, David and Fischer, Quirin and Hashme, Shariq and Hesse, Chris and others},
  journal = {arXiv preprint arXiv:1912.06680},
  year    = {2019}
}

@inproceedings{baker2022vpt,
  title     = {Video {PreTraining} ({VPT}): Learning to Act by Watching Unlabeled Online Videos},
  author    = {Baker, Bowen and Akkaya, Ilge and Zhokhov, Peter and Huizinga, Joost and Tang, Jie and Ecoffet, Adrien and Houghton, Brandon and Sampedro, Raul and Clune, Jeff},
  booktitle = {Advances in Neural Information Processing Systems (NeurIPS)},
  year      = {2022},
  note      = {arXiv:2206.11795}
}

@article{hafner2025dreamerv3,
  title     = {Mastering diverse control tasks through world models},
  author    = {Hafner, Danijar and Pasukonis, Jurgis and Ba, Jimmy and Lillicrap, Timothy},
  journal   = {Nature},
  year      = {2025},
  publisher = {Nature Publishing Group},
  doi       = {10.1038/s41586-025-08744-2}
}

@article{hafner2021crafter,
  title   = {Benchmarking the Spectrum of Agent Capabilities},
  author  = {Hafner, Danijar},
  journal = {arXiv preprint arXiv:2109.06780},
  year    = {2021}
}

@article{sima2024,
  title   = {Scaling Instructable Agents Across Many Simulated Worlds},
  author  = {{SIMA Team} and Raad, Maria Abi and Ahuja, Arun and Barros, Catarina and Besse, Frederic and Bolt, Andrew and Bolton, Adrian and Brownfield, Bethanie and Buttimore, Gavin and Cant, Max and others},
  journal = {arXiv preprint arXiv:2404.10179},
  year    = {2024}
}

@article{sestini2022rlgame,
  title   = {Towards Informed Design and Validation Assistance in Computer Games Using Imitation Learning},
  author  = {Sestini, Alessandro and Bergdahl, Joakim and Tollmar, Konrad and Bagdanov, Andrew D. and Gissl{\'e}n, Linus},
  journal = {arXiv preprint arXiv:2208.07811},
  year    = {2022}
}

@article{juliani2020unityml,
  title   = {Unity: A General Platform for Intelligent Agents},
  author  = {Juliani, Arthur and Berges, Vincent-Pierre and Teng, Ervin and Cohen, Andrew and Harper, Jonathan and Elion, Chris and Goy, Chris and Gao, Yuan and Henry, Hunter and Mattar, Marwan and Lange, Danny},
  journal = {arXiv preprint arXiv:1809.02627},
  year    = {2020}
}

@article{peng2017deeploco,
  title   = {{DeepLoco}: Dynamic Locomotion Skills Using Hierarchical Deep Reinforcement Learning},
  author  = {Peng, Xue Bin and Berseth, Glen and Yin, KangKang and van de Panne, Michiel},
  journal = {ACM Transactions on Graphics (Proc.\ SIGGRAPH)},
  volume  = {36},
  number  = {4},
  year    = {2017},
  doi     = {10.1145/3072959.3073602}
}

@article{peng2018deepmimic,
  title   = {{DeepMimic}: Example-Guided Deep Reinforcement Learning of Physics-Based Character Skills},
  author  = {Peng, Xue Bin and Abbeel, Pieter and Levine, Sergey and van de Panne, Michiel},
  journal = {ACM Transactions on Graphics (Proc.\ SIGGRAPH)},
  volume  = {37},
  number  = {4},
  year    = {2018},
  doi     = {10.1145/3197517.3201311}
}

@article{peng2021amp,
  title   = {{AMP}: Adversarial Motion Priors for Stylized Physics-Based Character Control},
  author  = {Peng, Xue Bin and Ma, Ze and Abbeel, Pieter and Levine, Sergey and Kanazawa, Angjoo},
  journal = {ACM Transactions on Graphics (Proc.\ SIGGRAPH)},
  volume  = {40},
  number  = {4},
  year    = {2021},
  doi     = {10.1145/3450626.3459670}
}

@article{peng2022ase,
  title   = {{ASE}: Large-Scale Reusable Adversarial Skill Embeddings for Physically Simulated Characters},
  author  = {Peng, Xue Bin and Guo, Yunrong and Halper, Lina and Levine, Sergey and Fidler, Sanja},
  journal = {ACM Transactions on Graphics (Proc.\ SIGGRAPH)},
  volume  = {41},
  number  = {4},
  year    = {2022},
  doi     = {10.1145/3528223.3530110}
}

@article{hwangbo2019learning,
  title   = {Learning agile and dynamic motor skills for legged robots},
  author  = {Hwangbo, Jemin and Lee, Joonho and Dosovitskiy, Alexey and Bellicoso, Dario and Tsounis, Vassilios and Koltun, Vladlen and Hutter, Marco},
  journal = {Science Robotics},
  volume  = {4},
  number  = {26},
  pages   = {eaau5872},
  year    = {2019},
  doi     = {10.1126/scirobotics.aau5872}
}

@inproceedings{siekmann2021periodic,
  title     = {Sim-to-Real Learning of All Common Bipedal Gaits via Periodic Reward Composition},
  author    = {Siekmann, Jonah and Godse, Yesh and Fern, Alan and Hurst, Jonathan},
  booktitle = {IEEE International Conference on Robotics and Automation (ICRA)},
  pages     = {7309--7315},
  year      = {2021},
  doi       = {10.1109/ICRA48506.2021.9561814}
}

@inproceedings{cheng2024extremeparkour,
  title     = {Extreme Parkour with Legged Robots},
  author    = {Cheng, Xuxin and Shi, Kexin and Agarwal, Ananye and Pathak, Deepak},
  booktitle = {IEEE International Conference on Robotics and Automation (ICRA)},
  year      = {2024},
  note      = {arXiv:2309.14341}
}

@article{tassa2018dmcontrol,
  title   = {{DeepMind} Control Suite},
  author  = {Tassa, Yuval and Doron, Yotam and Muldal, Alistair and Erez, Tom and Li, Yazhe and de Las Casas, Diego and Budden, David and Abdolmaleki, Abbas and Merel, Josh and Lefrancq, Andrew and Lillicrap, Timothy and Riedmiller, Martin},
  journal = {arXiv preprint arXiv:1801.00690},
  year    = {2018}
}

@inproceedings{makoviychuk2021isaacgym,
  title     = {{Isaac Gym}: High Performance {GPU}-Based Physics Simulation For Robot Learning},
  author    = {Makoviychuk, Viktor and Wawrzyniak, Lukasz and Guo, Yunrong and Lu, Michelle and Storey, Kier and Macklin, Miles and Hoeller, David and Rudin, Nikita and Allshire, Arthur and Handa, Ankur and State, Gavriel},
  booktitle = {Advances in Neural Information Processing Systems Datasets and Benchmarks Track},
  year      = {2021},
  note      = {arXiv:2108.10470}
}

@misc{zakka2025mujocoplayground,
  title  = {{MuJoCo} Playground},
  author = {Zakka, Kevin and Tabanpour, Baruch and Liao, Qiayuan and Haiderbhai, Mustafa and Holt, Samuel and Luo, Jing Yuan and Allshire, Arthur and Frey, Erik and Sreenath, Koushil and Kahrs, Lueder A. and Sferrazza, Carmelo and Tassa, Yuval and Abbeel, Pieter},
  year   = {2025},
  note   = {Robotics: Science and Systems (RSS) 2025, Outstanding Demo Paper Award. arXiv:2502.08844}
}

@article{brockman2016openaigym,
  title   = {{OpenAI Gym}},
  author  = {Brockman, Greg and Cheung, Vicki and Pettersson, Ludwig and Schneider, Jonas and Schulman, John and Tang, Jie and Zaremba, Wojciech},
  journal = {arXiv preprint arXiv:1606.01540},
  year    = {2016}
}

@article{towers2024gymnasium,
  title   = {Gymnasium: A Standard Interface for Reinforcement Learning Environments},
  author  = {Towers, Mark and Kwiatkowski, Ariel and Terry, Jordan and Balis, John U. and De Cola, Gianluca and Deleu, Tristan and Goul{\~a}o, Manuel and Kallinteris, Andreas and Krimmel, Markus and KG, Arjun and others},
  journal = {arXiv preprint arXiv:2407.17032},
  year    = {2024}
}

@misc{embark2025arcraiders,
  title        = {{ARC Raiders}},
  author       = {{Embark Studios}},
  year         = {2025},
  howpublished = {Video game. Released 30 October 2025},
  url          = {https://www.arcraiders.com/}
}

@article{romeo2025speq,
  title   = {{SPEQ}: Offline Stabilization Phases for Efficient {Q}-Learning in High Update-To-Data Ratio Reinforcement Learning},
  author  = {Romeo, Carlo and Macaluso, Girolamo and Sestini, Alessandro and Bagdanov, Andrew D.},
  journal = {Reinforcement Learning Journal (Proc. RLC 2025)},
  year    = {2025},
  note    = {arXiv:2501.08669}
}

@article{SOPE,
  title   = {{SOPE}: Stabilizing Off-Policy Evaluation for Online {RL} with Prior Data},
  author  = {Romeo, Carlo and Macaluso, Girolamo and Sestini, Alessandro and Bagdanov, Andrew D.},
  journal = {arXiv preprint arXiv:2605.05863},
  year    = {2026}
}

@inproceedings{freeman2021brax,
  title     = {{Brax} -- A Differentiable Physics Engine for Large Scale Rigid Body Simulation},
  author    = {Freeman, C. Daniel and Frey, Erik and Raichuk, Anton and Girgin, Sertan and Mordatch, Igor and Bachem, Olivier},
  booktitle = {Advances in Neural Information Processing Systems Datasets and Benchmarks Track},
  year      = {2021},
  note      = {arXiv:2106.13281}
}

@inproceedings{hutter2016anymal,
  title     = {{ANYmal} -- A Highly Mobile and Dynamic Quadrupedal Robot},
  author    = {Hutter, Marco and Gehring, Christian and Jud, Dominic and Lauber, Andreas and Bellicoso, C. Dario and Tsounis, Vassilios and Hwangbo, Jemin and Bodie, Karen and Fankhauser, P{\'e}ter and Bloesch, Michael and Diethelm, Remo and Bachmann, Samuel and Melzer, Amir and Hoepflinger, Mark},
  booktitle = {IEEE/RSJ International Conference on Intelligent Robots and Systems (IROS)},
  pages     = {38--44},
  year      = {2016},
  doi       = {10.1109/IROS.2016.7758092}
}

@misc{unitree_go1,
  title  = {{Unitree Go1}},
  author = {{Unitree Robotics}},
  year   = {2021},
  note   = {\url{https://www.unitree.com/go1/}}
}

@misc{bostondynamics_spot,
  title  = {{Spot}: The Agile Mobile Robot},
  author = {{Boston Dynamics}},
  year   = {2024},
  note   = {\url{https://bostondynamics.com/products/spot/}}
}

@inproceedings{xie2018cassie,
  title     = {Feedback Control For {Cassie} With Deep Reinforcement Learning},
  author    = {Xie, Zhaoming and Berseth, Glen and Clary, Patrick and Hurst, Jonathan and van de Panne, Michiel},
  booktitle = {IEEE/RSJ International Conference on Intelligent Robots and Systems (IROS)},
  year      = {2018},
  doi       = {10.1109/IROS.2018.8593722}
}

@article{caluwaerts2023barkour,
  title   = {{Barkour}: Benchmarking Animal-level Agility with Quadruped Robots},
  author  = {Caluwaerts, Ken and Iscen, Atil and Kew, J. Chase and Yu, Wenhao and Zhang, Tingnan and Freeman, Daniel and Lee, Kuang-Huei and Lee, Lisa and Saliceti, Stefano and Zhuang, Vincent and others},
  journal = {arXiv preprint arXiv:2305.14654},
  year    = {2023}
}

@article{lee2020learning,
  title   = {Learning quadrupedal locomotion over challenging terrain},
  author  = {Lee, Joonho and Hwangbo, Jemin and Wellhausen, Lorenz and Koltun, Vladlen and Hutter, Marco},
  journal = {Science Robotics},
  volume  = {5},
  number  = {47},
  pages   = {eabc5986},
  year    = {2020},
  doi     = {10.1126/scirobotics.abc5986}
}

@article{miki2022learning,
  title   = {Learning robust perceptive locomotion for quadrupedal robots in the wild},
  author  = {Miki, Takahiro and Lee, Joonho and Hwangbo, Jemin and Wellhausen, Lorenz and Koltun, Vladlen and Hutter, Marco},
  journal = {Science Robotics},
  volume  = {7},
  number  = {62},
  pages   = {eabk2822},
  year    = {2022},
  doi     = {10.1126/scirobotics.abk2822}
}

@inproceedings{rudin2022walk,
  title     = {Learning to Walk in Minutes Using Massively Parallel Deep Reinforcement Learning},
  author    = {Rudin, Nikita and Hoeller, David and Reist, Philipp and Hutter, Marco},
  booktitle = {Proceedings of the 5th Conference on Robot Learning (CoRL)},
  series    = {Proceedings of Machine Learning Research},
  volume    = {164},
  pages     = {91--100},
  year      = {2022}
}

@inproceedings{peng2020imitating,
  title     = {Learning Agile Robotic Locomotion Skills by Imitating Animals},
  author    = {Peng, Xue Bin and Coumans, Erwin and Zhang, Tingnan and Lee, Tsang-Wei Edward and Tan, Jie and Levine, Sergey},
  booktitle = {Robotics: Science and Systems (RSS)},
  year      = {2020},
  doi       = {10.15607/RSS.2020.XVI.064}
}

@article{shao2022phase,
  title   = {Learning Free Gait Transition for Quadruped Robots via Phase-Guided Controller},
  author  = {Shao, Yecheng and Jin, Yongbin and Liu, Xianwei and He, Weiyan and Wang, Hongtao and Yang, Wei},
  journal = {IEEE Robotics and Automation Letters},
  volume  = {7},
  number  = {2},
  pages   = {1230--1237},
  year    = {2022},
  doi     = {10.1109/LRA.2021.3136645}
}

@inproceedings{margolis2023walk,
  title     = {Walk These Ways: Tuning Robot Control for Generalization with Multiplicity of Behavior},
  author    = {Margolis, Gabriel B. and Agrawal, Pulkit},
  booktitle = {Proceedings of the 6th Conference on Robot Learning (CoRL)},
  series    = {Proceedings of Machine Learning Research},
  volume    = {205},
  year      = {2023}
}

@article{ijspeert2008,
  title   = {Central pattern generators for locomotion control in animals and robots: A review},
  author  = {Ijspeert, Auke Jan},
  journal = {Neural Networks},
  volume  = {21},
  number  = {4},
  pages   = {642--653},
  year    = {2008},
  doi     = {10.1016/j.neunet.2008.03.014}
}

@inproceedings{iscen2018pmtg,
  title     = {Policies Modulating Trajectory Generators},
  author    = {Iscen, Atil and Caluwaerts, Ken and Tan, Jie and Zhang, Tingnan and Coumans, Erwin and Sindhwani, Vikas and Vanhoucke, Vincent},
  booktitle = {Conference on Robot Learning (CoRL)},
  series    = {Proceedings of Machine Learning Research},
  volume    = {87},
  pages     = {916--926},
  year      = {2018}
}

@article{bellegarda2022cpgrl,
  title   = {{CPG-RL}: Learning Central Pattern Generators for Quadruped Locomotion},
  author  = {Bellegarda, Guillaume and Ijspeert, Auke},
  journal = {IEEE Robotics and Automation Letters},
  volume  = {7},
  number  = {4},
  pages   = {12547--12554},
  year    = {2022},
  doi     = {10.1109/LRA.2022.3218167}
}

@inproceedings{bellegarda2024visualcpgrl,
  title     = {Visual {CPG-RL}: Learning Central Pattern Generators for Visually-Guided Quadruped Locomotion},
  author    = {Bellegarda, Guillaume and Shafiee, Milad and Ijspeert, Auke},
  booktitle = {IEEE International Conference on Robotics and Automation (ICRA)},
  pages     = {1420--1427},
  year      = {2024}
}

@inproceedings{shafiee2024manyquadrupeds,
  title     = {{ManyQuadrupeds}: Learning a Single Locomotion Policy for Diverse Quadruped Robots},
  author    = {Shafiee, Milad and Bellegarda, Guillaume and Ijspeert, Auke},
  booktitle = {IEEE International Conference on Robotics and Automation (ICRA)},
  year      = {2024},
  note      = {arXiv:2310.10486}
}

@inproceedings{zhang2023synloco,
  title     = {{SYNLOCO}: Synthesizing Central Pattern Generator with Reinforcement Learning for Quadruped Locomotion},
  author    = {Zhang, Xinyu and Xiao, Zhiyuan and Zhang, Qingrui and Pan, Wei},
  booktitle = {IEEE Conference on Decision and Control (CDC)},
  year      = {2024},
  note      = {arXiv:2310.06606. Authors corrected from earlier draft, which incorrectly attributed the paper to Bellegarda et al.}
}

@article{li2024aicpg,
  title   = {{AI-CPG}: Adaptive Imitated Central Pattern Generators for Bipedal Locomotion Learned Through Reinforced Reflex Neural Networks},
  author  = {Li, Guanda and Ijspeert, Auke and Hayashibe, Mitsuhiro},
  journal = {IEEE Robotics and Automation Letters},
  volume  = {9},
  number  = {6},
  pages   = {5190--5197},
  year    = {2024},
  doi     = {10.1109/LRA.2024.3388842}
}

@inproceedings{siekmann2021stairs,
  title     = {Blind Bipedal Stair Traversal via Sim-to-Real Reinforcement Learning},
  author    = {Siekmann, Jonah and Green, Kevin and Warila, John and Fern, Alan and Hurst, Jonathan},
  booktitle = {Robotics: Science and Systems (RSS)},
  year      = {2021},
  doi       = {10.15607/RSS.2021.XVII.061}
}

@inproceedings{todorov2012mujoco,
  title     = {{MuJoCo}: A physics engine for model-based control},
  author    = {Todorov, Emanuel and Erez, Tom and Tassa, Yuval},
  booktitle = {IEEE/RSJ International Conference on Intelligent Robots and Systems (IROS)},
  pages     = {5026--5033},
  year      = {2012},
  doi       = {10.1109/IROS.2012.6386109}
}

@inproceedings{mysore2021caps,
  title     = {Regularizing Action Policies for Smooth Control with Reinforcement Learning},
  author    = {Mysore, Siddharth and Mabsout, Bassel and Mancuso, Renato and Saenko, Kate},
  booktitle = {IEEE International Conference on Robotics and Automation (ICRA)},
  pages     = {1810--1816},
  year      = {2021},
  doi       = {10.1109/ICRA48506.2021.9561138}
}

@inproceedings{haarnoja2018soft,
  title     = {Soft Actor-Critic: Off-Policy Maximum Entropy Deep Reinforcement Learning with a Stochastic Actor},
  author    = {Haarnoja, Tuomas and Zhou, Aurick and Abbeel, Pieter and Levine, Sergey},
  booktitle = {Proceedings of the 35th International Conference on Machine Learning (ICML)},
  series    = {Proceedings of Machine Learning Research},
  volume    = {80},
  pages     = {1861--1870},
  year      = {2018}
}

@article{sacfd1,
  title   = {Mapless navigation for {UAVs} via reinforcement learning from demonstrations},
  author  = {Yang, Jiaqi and Lu, Songyi and Han, Miao and Li, Yuze and Ma, Yongqi and Lin, Zihao and Li, Hangxin},
  journal = {Science China Technological Sciences},
  volume  = {66},
  number  = {5},
  pages   = {1263--1270},
  year    = {2023},
  doi     = {10.1007/s11431-022-2292-3}
}
\bibliographystyle{rlj}

\end{document}